# Comparative Design Space Exploration of Dense and Semi-Dense SLAM

M. Zeeshan Zia, Luigi Nardi, Andrew Jack, Emanuele Vespa, Bruno Bodin, Paul H.J. Kelly, Andrew J. Davison

*Abstract*— SLAM has matured significantly over the past few years, and is beginning to appear in serious commercial products. While new SLAM systems are being proposed at every conference, evaluation is often restricted to qualitative visualizations or accuracy estimation against a ground truth. This is due to the lack of benchmarking methodologies which can holistically and quantitatively evaluate these systems. Further investigation at the level of individual kernels and parameter spaces of SLAM pipelines is non-existent, which is absolutely essential for systems research and integration. We extend the recently introduced SLAMBench framework to allow comparing two state-of-the-art SLAM pipelines, namely KinectFusion and LSD-SLAM, along the metrics of accuracy, energy consumption, and processing frame rate on two different hardware platforms, namely a desktop and an embedded device. We also analyze the pipelines at the level of individual kernels and explore their algorithmic and hardware design spaces for the first time, yielding valuable insights.

## I. INTRODUCTION

The past five years have seen an explosion in SLAM research, with major advances seen across various directions: increasingly richer scene reconstruction, maturing of visual-inertial navigation systems, scaling to large environments, dynamic scenes, and semantic SLAM. In fact, there have even been discussions on whether SLAM has already been solved, as "demonstrated" by a number of commercial products deploying SLAM in the real world, e.g. Google Project Tango, Microsoft Hololens, Qualcomm Vuforia, Dyson 360 Eye vacuum cleaners, and advanced driver assistance systems. To sustain this rate of innovation, and as SLAM moves towards being a mature module in real-world robots, the importance of holistically comparing SLAM pipelines along different metrics is becoming even more pronounced.

Unfortunately, there is little work done on comparing different SLAM pipelines, analyzing the breakdown into computational kernels or analyzing individual design parameters. Papers introducing new SLAM algorithms usually limit themselves to qualitative comparisons on a few video sequences, or at most give accuracy comparison in terms of localization error. In fact, given many real-world constraints that robots have to obey, SLAM is a multi-dimensional (accuracy, processing time, energy consumption) and multi-domain (algorithmic, compilation, hardware) optimization problem. While the many trade-offs involved mean that it is unlikely any SLAM pipeline is going to be superior to another at *all* points in the configuration or use-case space.

Unlike other areas of computer vision like object detection and tracking, the lack of benchmarking methodologies in SLAM have resulted in little work being done on input parameter exploration, or thorough comparison of different SLAM systems. Only recently papers have appeared which allow quantifying the accuracy of pipelines [14], [5], and benchmarking SLAM from a multi-objective optimization perspective [10]. It should be clear that for systems integration, such holistic and quantitative comparison is essential. For example, smartphones have a strict energy budget and small UAVs additionally need to adapt their operating point for the accuracy/energy consumption trade-off dynamically depending upon how cluttered the environment is and how much battery they have left. In this work, we leverage the recently proposed SLAMBench framework [10] to holistically evaluate two different state-of-the-art SLAM pipelines along the axes of accuracy, energy consumption, and speed, and evaluate their design spaces.

In summary, this paper makes the following contributions:

- Analysis of two state-of-the-art SLAM systems, namely KinectFusion [11] and LSD-SLAM [2], including breakdown of these pipelines into constituent kernels and classification into respective parallel patterns.
- Analysis of how input parameter spaces (algorithmic, hardware) affect the performance for these pipelines.
- Analysis of energy, accuracy, and frame rate tradeoffs for the two algorithms.
- Extend SLAMBench with LSD-SLAM [2], and TUM RGB-D dataset [14], to support competitive, multimodal optimization of different SLAM implementations w.r.t. quality of result under controlled conditions.

## II. RELATED WORK

Visual SLAM research has focused on sparse keypoint based front ends with three major paradigms for inference. These include approaches based on Kalman filters [1], particle filters [3], and bundle adjustment [7]. Strasdat et al [13] performed a systematic comparison between Kalman filtering (EKF) and bundle adjustment approaches, showing that having a large number of keypoints is more important for performing accurate SLAM than having a large number of frames. Since the computational requirements of bundle adjustment methods do not grow with the number of keypoints as severely as EKF-based approaches, bundle adjustment methods are superior in practice for SLAM back-ends.

M.Z. Zia, L. Nardi, A. Jack, E. Vespa, P.H.J. Kelly, and A.J. Davison are with the Department of Computing, Imperial College London, UK {zeeshan.zia, l.nardi, andrew.jack09, e.vespa14, p.kelly, ajd}@imperial.ac.uk
B. Bodin is with the School of Informatics, University of Edinburgh, UK bbodin@inf.ed.ac.uk

Unfortunately, few other studies exist comparing different SLAM modules in a systematic way as [13] does.

Recently, a new front-end for SLAM has been proposed from the computer vision community, embedded in a complete pipeline called LSD-SLAM [2]. This "direct" front-end utilizes edge pixels in the images and performs image-to-image alignment by whole image matching as opposed to extracting salient keypoints and performing matching on these sparse patches encoded by invariant descriptors. Utilizing a larger portion of data contained within each image, as opposed to discarding all but a sparse sampling of keypoints, is definitely a promising approach, even if it is not yet as mature as keypoint-based approaches [9].

With the advent of consumer RGB-D cameras, Kinect-Fusion [11] emerged as one of the first systems to perform dense SLAM in real-time. While the map representation here limits the scale of the 3D reconstruction, alternative methods and data structures have emerged that build on KinectFusion to allow significantly larger scale environments.

A quantitative analysis of the reconstruction accuracy achievable as a function of input voxel grid resolution has been performed in [16]. However, the analysis is restricted to just one algorithmic input parameter, evaluating only along one metric (accuracy), whereas we evaluate and compare across multiple input parameters and three metrics.

Recently two datasets for quantitative evaluation of the accuracy of hand-held six degree of freedom SLAM systems have been proposed: the TUM RGBD dataset [14] and the ICL-NUIM dataset [5]. Both these datasets provide ground truth camera trajectories, against real [14] and synthetic [5] indoor scenes recorded/rendered with an RGB-D camera.

A recent innovation in comparing SLAM approaches in a holistic way is the SLAMBench [10] framework, which enables fine-grained quantitative analysis of SLAM kernels and pipelines along multiple metrics, including ICL-NUIM [5] for accuracy estimation. SLAMBench includes implementations of KinectFusion in different languages across multiple hardware platforms, and provides instrumentation to measure accuracy, computational time, and energy consumption at a fine-grained level. In the present work, we extend SLAM-Bench with LSD-SLAM [2], integrate the TUM RGB-D dataset [14], and upgrade the energy measurement instrumentation to work with Intel (in addition to ARM) processors.

## III. SLAM PIPELINES

We briefly review the two pipelines and compare them.

### A. KinectFusion

KinectFusion [11] registers and fuses the stream of measured depth frames as the scene is explored from different viewpoints into a clean 3D geometric map. It normalizes each depth frame and applies a bilateral filter (*preprocess*), before computing a point cloud (with normals) for each pixel. Next, KinectFusion estimates (*track*) the new 3D pose of the camera by registering this point cloud against a synthetic rendering of current global map using a variant of iterative closest point (ICP). Once the new camera pose has been estimated, the corresponding depth map is fused into the current 3D reconstruction (*integrate*). KinectFusion utilizes a voxel grid as the data structure to represent the map, employing a truncated signed distance function (TSDF) to represent 3D surfaces.

### B. LSD-SLAM

LSD-SLAM [2] represents the map as a set of keyframes, while improving the depth estimates at each edge pixel (i.e. all those pixels which have sufficiently high intensity gradient). The pipeline comprises of four modules: *tracking*, *depth estimation*, *loop closure detection*, and *global optimization*. Given a new monocular image, the *tracking* module estimates a rigid transformation aligning the image against the current reference keyframe. The *depth estimation* thread either converts the frame into a new keyframe propagating the prior depth estimates to the new keyframe; or performs small-baseline stereo, probabilistically refining the current depth estimates. If the decision to choose a new keyframe is made, the current keyframe is added to the pose-graph in the *global optimization* module. The addition involves retrieving a few of the nearest keyframes already present in the map, estimating similarity transformations to those keyframes, and finally adding these transformations as edges in the pose graph. The module calls g2o [8] to perform global pose graph optimization. In parallel, a *loop closure detection* thread, also called *constraint search thread*, uses appearance-based matching to find large-scale matches between keyframes and obtains new constraints to be inserted into the graph.

### C. Qualitative differences between the pipelines

As it should already be clear from the previous sections, KinectFusion and LSD-SLAM have a number of differences. Comparing SLAM pipelines that are so different is interesting exactly because it allows us to examine a large variation on the performance metrics. In the following we make these qualitative differences explicit.

*Sensors: Depth vs Image Intensity.* The input sensors to both the pipelines are different. KinectFusion utilizes depth frames estimated by a special purpose system-on-chip embedded on the Kinect device. This causes significant savings on the computational front while the active sensing requires additional power expenditure. LSD-SLAM, on the other hand, has to accumulate depth information in a probabilistic framework from a number of frames, which is its most expensive block (Sect. V). Of course active sensing enables more precise depth estimation just from a single view. Unfortunately, our analysis does not take these sensor-level differences into account but, as we shall see, we are still able to gain useful insights.

*Dense vs semi-dense front-end.* A dense front-end utilizes all depth pixels in the input frames, whereas a semi-dense front-end uses only intensity pixels with strong edges. In principle, being able to use all the information in the image frame should provide better estimates, as opposed to discarding a large number of pixels, even if they are less-informative.

*Map representation.* The map is represented as a dense TSDF voxel grid in KinectFusion, whereas in LSD-SLAM it is represented by a set of keyframes and depth estimates for the high-gradient pixels in those frames. The voxel grid is wasteful, requiring empty regions of the space to be represented and processed, whereas keyframes encoded at the level of semi-dense edge pixels are much more compressed.

*Tracking.* In KinectFusion the camera is tracked against synthetic rendering of the TSDF voxel grid. The relative pose is estimated by running iterated closest point (ICP). On the other hand, LSD-SLAM maintains a current keyframe at all instances against which any incoming frame is tracked by direct image alignment. The raycasting operation required in KinectFusion follows a search pattern [10], and is one of the most expensive kernels in the pipeline, whereas image-to-image alignment is far cheaper computationally.

*TSDF averaging vs global optimization.* KinectFusion fuses data from incoming frames into the map representation by computing a running average on the aligned TSDF voxel grid. On the other hand, LSD-SLAM keeps adding new keyframes and performs global pose-graph optimization every now and then. Here TSDF integration is computationally expensive, whereas global pose optimization is cheap.

*Loop closing.* The keyframe-based map computes and stores rigid transformations between frames represented as edges in the pose-graph. This enables introducing new constraints provided by a loop closure detection module, into the graph, and subsequently optimizing to reach a self-consistent map. However, it is not clear how to perform map deformation to accommodate loop closures in a voxel grid representation.

## IV. IMPLEMENTATION

One contribution of the present paper is in significantly extending the SLAMBench framework [10] along multiple axes. We highlight important integration and modification decisions in the following, some of which also have an effect on the experimental evaluation presented in Section V. We will make these extensions public.

### A. LSD-SLAM integration

The publicly available implementation of LSD-SLAM relies upon the Robot Operating System library (ROS), which is primarily supported only on Ubuntu Linux, to access the camera stream. To make the SLAMBench framework less dependent on existing software, we start with an implementation [17] that removes this dependency.

This first integration challenge is to enforce that all frames be processed, as done in SLAMBench for KinectFusion, in a mode called *process-every-frame*, to ensure repeatability of experiments. Unfortunately, due to the multi-threaded nature of LSD-SLAM this is not trivial. Another related issue is that of achieving deterministic behaviour.

The challenge in enabling a *process-every-frame* mode within LSD-SLAM is to avoid dropping any frames during handovers between the asynchronous threads. Specifically, incoming frames are tracked against the current keyframe (*track* thread), and placed into a temporary buffer. Later, they are removed and used to update the keyframes' depth map (*depth estimation* thread). However, whilst frames are stored in the buffer, the current keyframe can be replaced. 'Old' keyframes are stored in the pose graph, and cannot be updated. Therefore, frames tracked against these old keyframes are dropped as they are no longer useful. Between repeated executions of the algorithm, over a fixed dataset, the dropped frames vary slightly due to the asynchronous nature of the threads, hence not all frames are processed equally; this also contributes to non-deterministic behaviour.

In the new *process-every-frame* mode, we solve this problem by waiting (not introducing a new frame into the system), until the current frame has been processed by the depth mapping thread.

Another prior limitation which would have restricted our intended use of LSD-SLAM integration is the severe non-deterministic behavior of the system. This behavior results in variation of as much as $0.5$ cm in absolute trajectory error (ATE, sometimes called mean absolute error MAE [14]) upon multiple runs with *TUM RGB-D fr2/desk* sequence. The major source of non-determinism lies in the *loop closure detection* block: pre-emption in the thread, and the use of signals. We specify a pseudo-code version of the original implementation in Algorithm 1 which can exit in three ways: the new keyframe signal is fired due to a real new keyframe, a spurious wake-up due to implementation details of the *SLEEP* function used, or the timer timing out. These factors combined mean that some keyframes can be randomly considered for constraint-search more than others. Further, the order in which keyframes and constraints are added to the pose graph, used by *g2o*, affects the optimization results even if they are all added before the next optimization search.

In order to keep most of the existing behavior, our solution is three fold. Firstly, within the loop in Algorithm 1 we only consider adding constraints when a new keyframe is added. Before LSD-SLAM reports the poses of all frames, a final constraint search (loop closure detection) and optimization is performed, for each frame, ensuring no constraints are missed. Secondly, we sort the constraints before adding them to the pose graph. Finally, we only perform pose graph optimization after the constraints have been added. The result is that we get the exact same ATE upon multiple runs of LSD-SLAM on a given machine.

Yet another new feature is to make algorithmic input parameters accessible from command line.

---

**Algorithm 1** Loop Closure Detection thread

1: **procedure** CONSTRAINTSEARCHTHREAD()
2:     **while** keepRunning **do**
3:         **if** new keyframe **then**
4:             FINDANDADDCONSTRAINTS($keyframe$)
5:         **else**
6:             $keyframe \leftarrow$ random $keyframe$ from graph
7:             FINDANDADDCONSTRAINTS($keyframe$)
8:             SLEEP(500ms)
9:         **end if**
10:    **end while**
11: **end procedure**

## B. TUM RGB-D dataset

In addition to LSD-SLAM, we also introduce a real scene dataset [14] into SLAMBench [10]. Although SLAMBench already ships with the ICL-NUIM [5] dataset, it is limited to synthetic scenes. The depth frames, with artificial noise, provided with the dataset appear quite reasonable, however the RGB images clearly lack realism. We also show the limitations of the synthetic dataset quantitatively in Section V. On the other hand, the TUM RGB-D dataset [14] has been captured with a Kinect camera, covering a larger variety of scenes and uses a motion capture system to provide ground truth camera trajectory.

The format and methodology for using the TUM RGB-D dataset [14] is different from ICL-NUIM [5]. The measurements, namely RGB, depth, and location are captured asynchronously in [14] which means additional glue code is needed for finding the closest corresponding location measurements for each frame. Secondly, since the origin is arbitrary in [14], a scale-aware variant of [6] is used to align the two trajectories (since monocular visual SLAM cannot provide true scale). We use this method for all ATE evaluations in this paper. Finally, converting this format to the raw file format of SLAMBench is non-trivial; we describe the details in the documentation.

## C. Energy measurement for Intel

SLAMBench [10] provides software modules to access the hardware energy counters on the ODROID board. However there are no provisions to measure energy for other important devices such as those from Intel or Nvidia. The Running Average Power Limit (RAPL) is a feature of Intel processors, designed with the aim of managing power usage. Energy usage, amongst other characteristics is provided through the Model Specific Registers (MSR). We integrate the Performance Application Programming Interface (PAPI) [15] for accessing these registers on Intel, while enabling a consistent interface to use performance measurement hardware provided by a number of vendors via the PAPI components. The Intel MSR registers provide three energy readings: package, PP0, PP1. The package measurement is for the whole CPU unit, whereas PP0 is the core components and PP1 is the non-core. We use the package measurement (PP0) in our experiments in Section V.

## V. EVALUATION

### A. Protocol and Setup

We experiment with two hardware platforms: one is a desktop processor and another is a state-of-the-art embedded device, to allow benchmarking for two very different families of use cases. The desktop platform has an Intel i7-4770 Haswell processor with 4 CPU cores, operating at 3.4 GHz, and running Ubuntu 14.04 (kernel 3.13.0). The embedded platform is an ODROID (XU3) with the Exynos 5422 SoC from ARM. It comprises of 4 Cortex-A15 "big" cores operating at 1.8 GHz and 4 Cortex-A7 "little" cores operating at 1.3 GHz, as well as a Mali GPU (which we do not use), and runs Ubuntu 14.04 (kernel 3.10.58).

We exclusively use an OpenMP implementation of Kinect-Fusion [10] to allow a fair comparison with LSD-SLAM which does not use the GPU. This is a realistic operating environment, since for many practical use-cases of SLAM such as augmented reality, the GPU is busy doing graphics rendering. We operate in the *process-every-frame* mode unless otherwise specified. As default setup for KinectFusion, we use volume size and resolution of $9.6^3 m^3$, 512 respectively. For LSD-SLAM we use the default parameters, except for the keyframe selection variables: 'KFUsageWeight', 'KFDistWeight', both of value 5.0. We enable OpenFABMap and disable keyframe reactivation.

In reporting the results of the algorithmic design space exploration, we restrict ourselves to parameters which have significant impact or show surprising behavior.

### B. Holistic Comparison of KinectFusion and LSD-SLAM

We perform a holistic comparison of KinectFusion and LSD-SLAM, along our three metrics, while running on a synthetic sequence (ICL-NUIM Living Room 2) and a real sequence (TUM RGB-D fr2/xyz). As detailed in Table I, LSD-SLAM runs $4.5 - 6x$ faster than KinectFusion, while consuming $9 - 19x$ less energy. This together with the fact that KinectFusion requires additional energy for the active sensor and effectively additional computational time which is presently hidden within the SoC in the depth camera, implies that KinectFusion is significantly wasteful, specially if all that is needed is to perform precise tracking. On the other hand, a runtime of $0.20$ seconds per frame, means that even (the full) LSD-SLAM is far from real-time performance on a cutting-edge consumer mobile device.

Another interesting observation is that while the computation time per frame is constant on a given platform for LSD-SLAM ($0.03s$ for Desktop and $0.20s$ for ODROID), the energy consumption is greater by almost $50\%$ for the real sequence compared to the synthetic sequence. This highlights the importance of explicitly benchmarking energy consumption, as it is not just proportional to computation time as is often implicitly assumed. Unsurprisingly, the energy consumption values for desktop processor are more than twice those for the ODROID SoC, which is optimized for mobile applications.

Yet another observation is that despite the same code being run on the two platforms (for both KinectFusion and LSD-SLAM), we do not get the same ATE. This is due to different compilers, operating systems, and hardware being used, *e.g.* different floating point approximation used.

Further, we perform a small experiment on hardware design space exploration with ODROID, and tabulate the results in Table II. The experiments involve turning on either the four "big" A15 cores or the four "little" A7 cores, or all eight cores together, while distributing the computation across these cores. Here we gain insights into exploiting heterogeneous computing platforms for computer vision, and observe the different character of the two pipelines. First, we notice that as expected the average computation time per frame generally goes down as the number and/or speed of cores increases. On the other hand, since the energy consumption of A7 cores is far smaller than on the A15 cores, distributing the workload between A7 and A15 cores results in less energy consumption than turning off the A7 cores and doing all computation on A15 cores. In a similar vein, we consistently notice that the relationship between energy expenditure and speed is super-linear: while A15 cores execute KinectFusion twice as fast as A7 cores, they require $2.9 - 4.3x$ energy to achieve this speedup. Similarly, A15 cores execute LSD-SLAM with a speedup of only $1.2 - 1.4x$ relative to A7 cores, but need $2.5 - 2.7x$ the energy. The lesson learnt is that for both these algorithms, it is better to have a larger number of low-power less-complex computational cores like A7, than more expensive and complex computational cores like A15. This insight should be useful for a system designer choosing a computational platform which will run a SLAM system. A SLAM-specific observation is that KinectFusion scales better than LSD-SLAM (with the number of cores). This is because while the KinectFusion pipeline has very little task parallelism, which means it is essentially a set of sequential modules, there is a lot of data parallelism inside those modules. On the contrary, LSD-SLAM is the opposite, having a certain degree of task parallelism allowing to distribute its workload across four parallel threads, but little data parallelism.

| Platform | Seq. | Time/frame (s) | | ATE (cm) | | Energy/frame (J) | |
|---|---|---|---|---|---|---|---|
| | | KF | LSD | KF | LSD | KF | LSD |
| *Desktop* | Syn. | 0.18 | 0.03 | 1.36 | 4.44 | 12.51 | 0.80 |
| *Desktop* | Real | 0.15 | 0.03 | 2.62 | 0.99 | 10.62 | 1.21 |
| *ODROID* | Syn. | 0.89 | 0.20 | 1.35 | 4.37 | 4.90 | 0.38 |
| *ODROID* | Real | 0.93 | 0.20 | 2.62 | 1.14 | 4.99 | 0.50 |

TABLE I: Holistic comparison table.

| Seq. | Hardware | Time/frame (s) | | ATE (cm) | | Energy/frame (J) | |
|---|---|---|---|---|---|---|---|
| | | KF | LSD | KF | LSD | KF | LSD |
| Syn. | A7 + A15 | 0.88 | 0.20 | 2.04 | 4.37 | 4.94 | 0.38 |
| | A15 | 1.10 | 0.24 | 2.04 | 4.39 | 5.60 | 0.46 |
| | A7 | 2.27 | 0.28 | 2.04 | 4.39 | 1.91 | 0.17 |
| Real | A7 + A15 | 0.93 | 0.20 | 2.62 | 0.97 | 4.99 | 0.47 |
| | A15 | 0.84 | 0.26 | 2.61 | 1.02 | 4.80 | 0.55 |
| | A7 | 1.90 | 0.36 | 2.62 | 1.00 | 1.11 | 0.22 |

TABLE II: Hardware design space exploration.

| Major kernels | Block | Pattern | Percent |
|---|---|---|---|
| *Convert mm to meters* | Preprocess | Gather | 0% |
| *Bilateral Filter* | Preprocess | Stencil | 4% |
| *Half Sample* | Track | Stencil | 0% |
| *Depth to Vertex* | | Map | 0% |
| *Vertex to Normal* | | Stencil | 0% |
| *Track* | | Map/Gather | 2% |
| *Reduce* | | Reduction | 2% |
| *Solve* | | Sequential | 0% |
| *Integrate* | Integrate | Map/Gather | 73% |
| *Raycast* | Raycast | Search/Stencil | 17% |

TABLE III: KinectFusion kernel classification and timings on desktop, TUM RGB-D fr2/xyz.

*C. Kernel-level breakdown of KinectFusion and LSD-SLAM*

As discussed in Sect. III-C, KinectFusion and LSD-SLAM have a number of dissimilarities. These dissimilarities make it difficult to compare the computational requirements of the two pipelines at the block-level. Still an analysis of the constituent kernels provides insights into the computational behavior of the pipelines. Further, we classify the kernels of LSD-SLAM into parallel patterns, as done in [10] for KinectFusion (which we repeat here). This classification coupled with the computational time taken by each kernel should enable systems researchers to design accelerators which work for a variety of SLAM pipelines.

We perform kernel-level profiling of KinectFusion running on the *desktop* machine. The percentage of time required by each kernel, together with classification into parallel patterns [10] is listed in Table III. As mentioned earlier, the blocks are executed in sequential order, but most kernels use OpenMP pragmas to exploit multiple cores to operate on parallel data. As opposed to the similar table in [10], we disable the graphical visualization kernels, and use the OpenMP implementation instead of the strictly sequential C++ implementation as it is more relevant to the multi-core case. The two most compute intensive kernels, by far, turn out to be the *integrate* and *raycast* kernels which take 73% and 17% of the computational time, respectively.

We further perform kernel-level profiling of LSD-SLAM also running on *desktop* in Table IV. As opposed to Kinect-Fusion, here the four blocks are running in parallel as separate threads. We neglect the percentage of time spent in the glue logic that lets these kernels communicate. However we list the total time per thread for running the *TUM RGB-D fr2/desk* sequence. We observe that the *depth estimation* thread takes the greatest amount of total compute time (48 seconds), which can be seen as analogous to the *integrate* block of KinectFusion which is also the most expensive. It should be noted that within the *depth* block, we avoid including the time required for creating new keyframes because it is negligible. Specifically, over the whole sequence the time required for creating a new keyframe is less than 4% (2 sec) of the time spent on updating the depth map. The second most expensive thread is *track* (34 seconds), which can be seen as doing the job of *track* plus *raycast* in KinectFusion. Surprisingly, the *global optimization* thread is the cheapest.

Together the Tables III and IV highlight that *Map/Gather* is the most important parallel pattern for SLAM, followed by *Stencil*, since kernels following this pattern of computation take the greatest amount of processing time. While traditionally domain-specific languages (DSLs) for image processing [12] have focused on stencil and reduce patterns exclusively, the observation from the preceding analysis indicates that including native support for Map/Gather patterns together with Stencil and Reduce in DSLs and/or hardware can enable fast and low energy implementations of SLAM pipelines.

*D. Analysis of KinectFusion parameters and metrics*

We analyze the dependence of localization accuracy and processing time on the voxel grid resolution into which the 3D data from individual frames is integrated, as well as the impact of imposing a threshold on ICP residual for tracking.

| Thread name | Major kernels | Description | Pattern | Percent |
|---|---|---|---|---|
| Tracking (vectorized) | *Calc. Residuals* | ⎫ Calculate components of the Levenberg–Marquardt (LM) algorithm | Map | 72% |
| | *Calc. Weights and Residuals* | | Map | 4% |
| | *Calc. Jacobians* | ⎭ | Map-Reduce | 9% |
| | *Solve* | Evaluate the LM algorithm given the above calculations | External | 0% |
| *Total* | | | | **34 s** |
| Depth | *Stereo Line Search* | Epipolar line search | Map | 43% |
| | *Fill Holes* | Increase density of depth map | Stencil | 20% |
| | *Regularize Depth Map* | Denoise the depth map | Stencil | 28% |
| | *Copy Depth Map to Frame* | Implementation specific overhead | Map | 6% |
| *Total* | | | | **48 s** |
| Constraint Search | *Find Euclidean Overlaps* | Get neighbour frames from graph, to insert new constraints | Search | 6% |
| | *Filter and Sorting* | Remove less optimal frames from results | Map | 4% |
| | *Calc. Residuals* | ⎫ Calculate components of the Levenberg—Marquardt (LM) algorithm between keyframe and neighbour frames | Map | 71% |
| | *Calc. Weights and Residuals* | | Reduce | 7% |
| | *Calc. Jacobian Matrix* | ⎭ | External | 12% |
| *Total* | | | | **19 s** |
| Optimization | *g2o Call* | Run iterations of global optimization | External | 99% |
| | *Update Graph* | Incorporate improvements from g2o into graph | Map | 1% |
| *Total* | | | | **3 s** |

TABLE IV: LSD-SLAM kernel classification and timings on desktop, TUM RGB-D fr2/desk

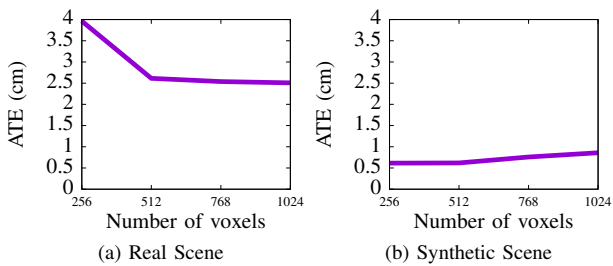

Fig. 1: Varying the voxel resolution in KinectFusion with default parameters on desktop, operating on the sequence: (a) TUM RGB-D fr2/xyz, (b) ICL-NUIM Living Room trajectory 1.

Further, we explore the effect of scene on accuracy and time.

We evaluate the impact of *voxel grid resolution* on the *ATE*, by applying KinectFusion to two sequences: ICL-NUIM Living Room Trajectory 1 and TUM RGB-D fr2/xyz while sweeping the parameter *volume resolution* over the settings $256^3, 512^3, 768^3, 1024^3$. We observe Figure 1(a) agreeing with the expected relationship of reducing *ATE* (going down from almost 4 cm to 2.5 cm) as volume resolution increases. However we also see the surprising result in Figure 1(b) that *ATE* can stay constant or even get slightly worse upon increasing voxel grid resolution; more than 0.2 cm in this case. We attribute this inconsistency to the complex relationship of *tracking* with noise of the 3D reconstruction. While the reconstruction improves qualitatively as seen in a visualization with higher resolution voxel grid, the representation also becomes noisier as depth samples are averaged into smaller voxels, implying a higher level of noise per voxel. In the future, we plan to incorporate novel metrics that evaluate reconstruction quality alongside ATE.

We further analyze the kernel timing distribution (as percentage of the total time) varying the *volume resolution* again in steps of $256^3, 512^3, 768^3, 1024^3$, plotted in Figure 2(a). We observe that the *integrate* stage in particular strongly depends upon the resolution of reconstruction, increasing in terms of compute load as the resolution increases. This is because of the need to traverse each voxel in the grid in the averaging step, whereas most other KinectFusion kernels have a sub-linear dependence on the number of voxels.

Another input parameter that is part of KinectFusion in the SLAMBench framework [10] is a threshold on error residuals obtained from each iteration of ICP for the *track* block. This threshold is used as an exit condition for ICP for a given instance of track kernel execution, together with a fixed number of maximum iterations. We found that the ATE is insensitive (in fact practically independent) to the ICP threshold chosen, despite sweeping this parameter over a range of five orders of magnitude, from $1 \times 10^{-6}$ to $1 \times 10^{-1}$.

We also analyzed the kernel timing distribution (as percentage of the total time) vs. scene geometry and camera trajectory: over two synthetic sequences (ICL-NUIM Living Room trajectory 1 and 2) and one real sequence (TUM RGB-D fr2/xyz), for fixed parameter settings. We find that the distribution stays fixed and essentially similar to Figure 2(a). We see that this is not the case with LSD-SLAM, where the computation time significantly depends upon the scene (the number of edge pixels).

*E. Analysis of LSD-SLAM parameters and metrics*

We perform design space exploration for the algorithmic parameters of LSD-SLAM, as well as an important hardware parameter namely the processor clock frequency. We also visualize the distribution of ATEs across different sequences.

We start by evaluating the *minimum pixel gradient* parameter, which specifies a threshold on the gradient magnitude. In LSD-SLAM, all pixels with an image gradient of magnitude smaller than this threshold are discarded. Thus a higher gradient threshold implies fewer edge pixels being used for alignment and depth computation. Figure 2(b) plots this parameter against ATE and FPS for two sequences. The curves terminate where the tracking fails for a single frame. As this threshold is increased and each image is left with fewer edge pixels, the frame rate increases from approximately 20 fps to 60 fps almost evenly for both sequences - with TUM RGB-D fr2/desk terminating earlier because there are not enough edge pixels left to allow reliable tracking. On the other hand, the ATE varies far less over most of the parameter sweep. The ATE however sharply improves for very low values of the threshold (for *fr2/desk*), which can be seen as the denoising effect of this parameter.

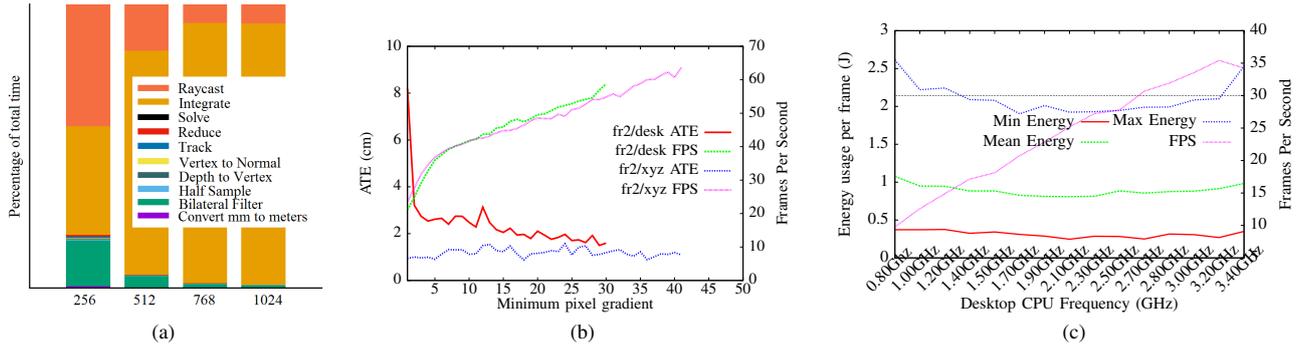

Fig. 2: (a) KinectFusion kernel timings as a percentage of the total time vs. the number of voxels (otherwise default parameters), running on desktop, on the ICL-NUIM Living Room Trajectory 1 (b) ATE and FPS using LSD-SLAM running on TUM RGB-D fr2/xyz and fr2/desk sequences sweeping minimum gradient threshold, running on desktop, (c) FPS and energy per frame of LSD-SLAM whilst changing CPU frequency, on desktop, using TUM RGB-D fr2/xyz.

Two other important algorithmic parameters control which frames become keyframes. The first one of these is the *keyframe to frame distance* which defines how often new keyframes are created, depending upon the Euclidean distance to the current keyframe; with larger value implying *more* keyframes. The second parameter is *keyframe to frame appearance similarity* which defines how often new keyframes are created based on the visual overlap with the current keyframe; again large values imply *more* keyframes Figure 3 visualizes the results of a two dimensional design space exploration within the range of reasonable parameter values [2]. Our first observation looking at the plots is that both accuracy and speed are fairly insensitive to the choice of these two parameters, with large regions of the plots having essentially the same color. However on the top-right of Figure 3(a) we start seeing slightly worse ATEs, as a result of too many keyframes, since there are not enough frames tracked against a given keyframe to propagate good estimates of depth. Similarly in Figure 3(b) we notice that the frame rate is relatively more affected by the appearance similarity parameter, which causes it to change expectedly, i.e. become slower for higher number of keyframes.

So far we have restricted ourselves to the *process-every-frame* mode, as it allows us to (partially) decorrelate the results of the algorithm from the scene and hardware/software implementation. However, it is still instructive to look at the performance of the system when the rate of incoming frames from the camera is fixed. This gives us insights into what happens as the incoming video frame rate exceeds the maximum processing frame rate. Figure 4 plots the ATE and energy consumption per frame for both platforms. Since the processing frame rate for desktop processor is greater than 30 FPS (Table I), no frames are dropped and essentially all frames are processed for all the input frame rates, yielding a fixed ATE. However, for the ODROID processor, which can process at maximum 5 fps (Table I), we notice a dramatic degradation of accuracy soon after 5 fps, with the ATE going from 2 cm to 15 cm. Unsurprisingly, energy consumption for ODROID also exhibits a linear dependency on the input frame rate, whereas that for the relatively wasteful desktop processor remains constant due to not reaching peak performance for this architecture.

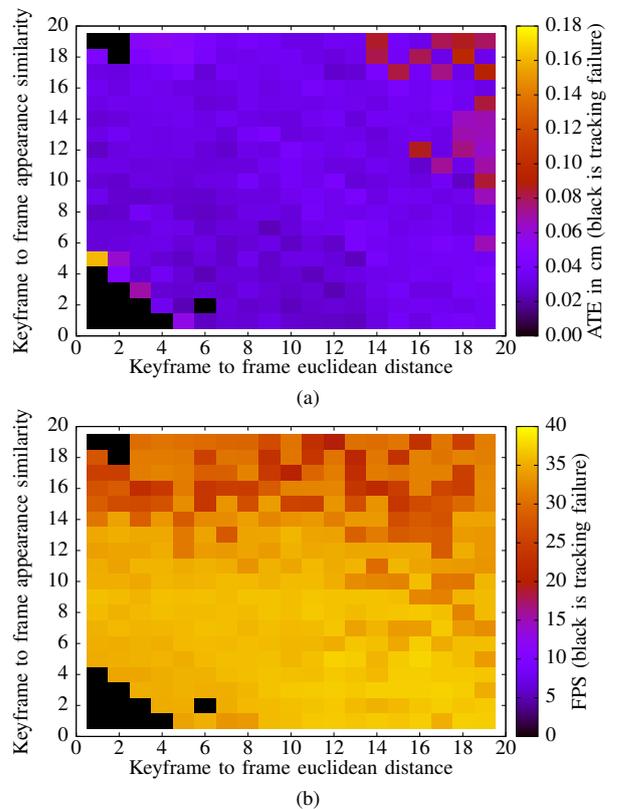

Fig. 3: 2D design space exploration of appearance similarity and distance between frames in LSD-SLAM using TUM RGB-D fr2/desk on Desktop with otherwise default parameters. The heatmap represents: (a) ATE (cm), (b) FPS.

As an important hardware parameter we sweep the desktop CPU frequency and explore the effect on energy consumption

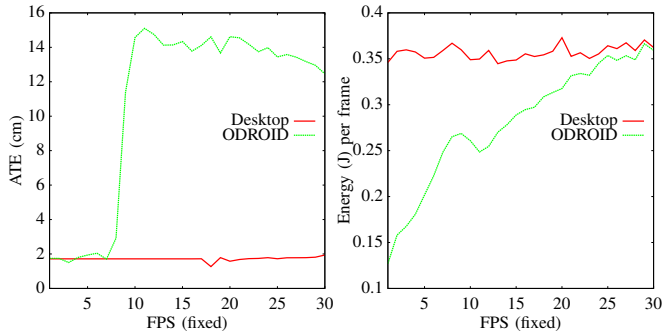

Fig. 4: Varying the input frame-rate in LSD-SLAM under the TUM RGB-D fr2/xyz dataset without using process-every-frame mode. Shows ATE and energy under Desktop and ODROID.

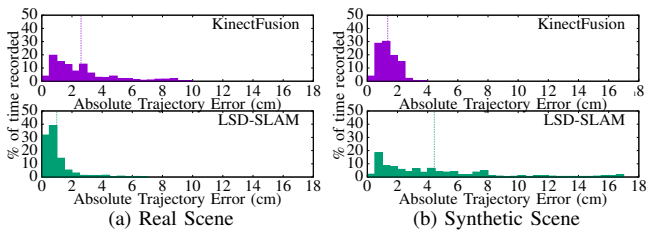

Fig. 5: Distribution of ATEs using KinectFusion and LSD-SLAM, run with default parameters on Desktop. MAE is highlighted. (a) TUM RGB-D fr2/xyz (b) ICL-NUIM Living Room Trajectory

per frame and frame rate in Figure 2(c). As in the previous experiment, we notice that the energy consumption remains fixed. However we also observe the frame rate to have a linear relationship with CPU frequency. This means that, for LSD-SLAM, running desktop-grade processors at the maximum clock frequency yields best performance (speed) while requiring the same energy.

*F. Dataset issues in SLAM*

We encounter inconsistent behavior of both the pipelines over the synthetic and the real datasets in terms of accuracy. Tables I and II already showed KinectFusion performing better than LSD-SLAM on the synthetic dataset; and opposite results on the real dataset. Looking at these results in finer detail, we plot the distribution of ATEs for both pipelines on a synthetic sequence in Figure 5(a) and on a real sequence 5(b). We observe KinectFusion not only outperforming LSD-SLAM on the synthetic sequence (in terms of mean error), but also the distribution of ATE across frames is tighter - which implies that fewer frames nearly fail tracking. On the real sequence we get the exact opposite behavior, with KinectFusion having a worse accuracy and greater variance. We attribute these contrasting observations to the shortcomings of the synthetic dataset, particularly the lack of realistic texture in the RGB images. Unfortunately, so far only synthetic datasets provide ground truth geometry together with camera trajectory. Laser scanning or other offline reconstruction method have not been used to provide that functionality in realistic datasets.

## VI. CONCLUSION

We exploit our extensions to SLAMBench [10] to analyze and contrast two state-of-the-art SLAM pipelines. We perform holistic comparison of the two pipelines along energy, accuracy, and speed across two hardware platforms: a desktop processor from Intel and a high-end embedded device from ARM. Further, we profile the kernel-level computational characteristics and classify the kernels into parallel design patterns. We also explore the algorithmic and hardware design spaces, and gain further insights into the behavior of these pipelines. This analysis should be of immense value for system-level design and integration, and the software would prove a valuable tool enabling performance optimization analysis for building high-performance SLAM systems. We plan to work on a more systematic exploration of motion and scenes with a new dataset [4].

## ACKNOWLEDGMENTS

We acknowledge funding by the EPSRC grant PAMELA EP/K008730/1. We thank Andy Nisbet and John Mawer, members of the PAMELA project at the University of Manchester, for developing the C++ and OpenMP implementations of KFusion part of SLAMBench used in this paper and for the software infrastructure to measure energy on the Intel and ARM platforms. We thank Jacob Engel for feedback on LSD-SLAM, and our steering group for discussions.